\documentclass[runningheads]{llncs}
\usepackage{color}
\usepackage[table]{xcolor}
\usepackage{graphicx}
\usepackage{multirow}
\usepackage{makecell}
\usepackage{booktabs}
\usepackage{amsmath}
\usepackage{bbm}
\usepackage[misc]{ifsym}
\usepackage{threeparttable}
\usepackage{cite}
\usepackage{amsmath,amssymb,amsfonts}
\usepackage{algorithmic}
\usepackage{textcomp}
\newcommand{\tabincell}[2]{\begin{tabular}{@{}#1@{}}#2\end{tabular}} 
\usepackage[colorlinks, citecolor=blue, anchorcolor=blue, linkcolor=blue]{hyperref}
\newcommand{\orcid}[1]{\href{https://orcid.org/#1}{\includegraphics[width=9pt]{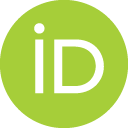}}}
\begin{document}
\title{Adversarial Learning-based Stance Classifier for COVID-19-related Health Policies}
\titlerunning{Stance Detection for COVID-19-related Health Policies}
\author{Feng Xie$^{*}$\orcid{0000-0003-3944-236X}\and
Zhong Zhang$^{*}$ \and
Xuechen Zhao \and
Haiyang Wang \and
Jiaying Zou \and\\
Lei Tian \and
Bin Zhou$^{(\textrm{\Letter})}$ \and
Yusong Tan}
\authorrunning{F. Xie et al.}
%
\institute{College of Computer, National University of Defense Technology
\email{\{xiefeng,zhangzhong,zhaoxuechen,wanghaiyang19,zoujiaying20\\leitian129,binzhou,ystan\}@nudt.edu.cn}}
\maketitle              
\begin{abstract}
\let\thefootnote\relax\footnotetext{* means equal contributions, \Letter {} means corresponding author.}

The ongoing COVID-19 pandemic has caused immeasurable losses for people worldwide. To contain the spread of the virus and further alleviate the crisis, various health policies (e.g., stay-at-home orders) have been issued which spark heated discussions as users turn to share their attitudes on social media. In this paper, we consider a  more realistic scenario on stance detection (i.e., cross-target and zero-shot settings) for the pandemic and propose an adversarial learning-based stance classifier to automatically identify the public’s attitudes toward COVID-19-related health policies. Specifically, we adopt adversarial learning that allows the model to train on a large amount of labeled data and capture transferable knowledge from source topics, so as to enable generalize to the emerging health policies with sparse labeled data. To further enhance the model's deeper understanding, we incorporate policy descriptions as external knowledge into the model. Meanwhile, a GeoEncoder is designed which encourages the model to capture unobserved background factors specified by each region and then represent them as non-text information. We evaluate the performance of a broad range of baselines on the stance detection task for COVID-19-related health policies, and experimental results show that our proposed method achieves state-of-the-art performance in both cross-target and zero-shot settings.

\end{abstract}
\section{Introduction}
\noindent The coronavirus disease 2019 (COVID-19) pandemic has brought serious challenges to human health, society, and the economy. To curb the spread of the virus and alleviate the crisis, policymakers and public authorities have imposed corresponding health policies  (e.g., stay-at-home orders, vaccination) based on the dynamics of the epidemic situation in response to complex virus challenges. The promulgation of these health policies has sparked discussion on the Internet, but the public's attitudes towards them vary. Stance detection is of great practical value as an effective tool for Internet public opinion monitoring, which detects the attitude (i.e., \textit{in favor of}, \textit{against}, or \textit{neutral}) of an opinionated text toward a pre-defined topic automatically \cite{wei2019modeling,allaway2021adversarial}, as is shown in Table \ref{tab:exp}.

\begin{table}[t]
    \renewcommand{\arraystretch}{1.4}
    \scriptsize
    \centering
    \caption{The examples of stance classification task. Given a tweet and the involved topic, the stance classifier is capable of detecting the \textbf{stance label} automatically.}
    \begin{tabular}{|lcl|}
        \hline
        \multicolumn{3}{|l|}{
        \begin{tabular}[l]{p{11.5cm}}
            \textbf{Example 1:} Don’t be selfish. Stay home, reduce the spread, and safe lives. If you have to go out, please wear a mask and gloves.
        \end{tabular}}
        \\
        \textbf{Topic:} Stay at Home Orders & & \quad\quad\quad\quad\quad\quad\quad\quad\quad\quad\quad\quad\quad\quad\quad\quad\quad\quad\textbf{Stance label:} \textcolor[RGB]{1,127,40}{Favor}\\
        \hline
        \multicolumn{3}{|l|}{
        \begin{tabular}[l]{p{11.5cm}}
            \textbf{Example 2:} There is no way in hell I would put a vaccine into my body that comes out under or is advocated by the trump regime. I trust COVID-19 more than I trust the trump regime.
        \end{tabular}}
        \\
        \textbf{Topic:} Vaccination & & \quad\quad\quad\quad\quad\quad\quad\quad\quad\quad\quad\quad\quad\quad\quad\quad\quad\quad\textbf{Stance label:} \textcolor{red}{Against}\\
        \hline
    \end{tabular}
    \label{tab:exp}
\end{table}

Nevertheless, for newly proposed health policies, the available labeled data is limited, and it is also infeasible to annotate a large amount of data in a short period of time. Existing studies adopt in-target\footnote{In this paper, we will use the terms: target and topic interchangeably.} setting for tracking the public's stances \cite{kuccuk2022sentiment,mather2021general}, that is, the training and testing phase are under the same health policy. However, they usually require adequate labeled data to achieve decent performance, which limits their application range. Therefore, it is imperative to propose a more practical and robust stance classifier to be applied to emerging health policies. The alternative strategies are to conduct cross-target \cite{xu2018cross,wei2019modeling} or zero-shot settings \cite{allaway2020vast,allaway2021adversarial}. Cross-target stance detection trains on one target and tests on a related target in a one-to-one way, while zero-shot stance detection aims to train on multiple targets with labeled data and evaluate on an unseen target. Both cross-target and zero-shot settings are more challenging principally because the language models may not be compatible between different targets \cite{xu2018cross}. Since COVID-19-related health policies are inherently correlated, adopting cross-target or zero-shot stance detection becomes viable and could provide more timely and useful guidance for administrative decision-making.

In this paper, we consider a  more realistic scenario on stance detection (i.e., cross-target and zero-shot settings) for the pandemic and propose an adversarial learning-based stance classifier to detect the public's attitudes toward COVID-19-related health policies. The domain adaptation technique is one of the best solutions applied to cross-target and zero-shot conditions \cite{wei2019modeling,allaway2021adversarial}, where domain-invariant information is responsible for ensuring the transferability across different domains (topics). We treat each health policy as a topic and model topic transfer as domain adaptation. Specifically, we embed a text and related policy descriptions jointly via Bidirectional Encoder Representations from Transformers (BERT) \cite{devlin2018bert}, and then we employ a feature separation module to extract and distinguish topic-specific and topic-invariant information. Moreover, since people's attitudes across different regions are influenced by unobserved background factors, such as cultural background, political ideologies, and regional epidemic situation (i.e., epidemiological context), a GeoEncoder is devised which encodes geographic signals as non-text representations to learn regional background information and thus improve the model's understanding. Following the success of adversarial learning for domain adaptation \cite{ganin2015unsupervised,cui2020gradually}, we integrate a topic discriminator into the model for adversarial training to better capture topic-invariant information, hence enhancing the transferability of applying it to the emerging health policies. Experiments conducted on COVID-19 stance datasets demonstrate that our proposed method achieves state-of-the-art performance in both cross-target and zero-shot settings against a broad range of baselines.

\section{Methodology}\label{sec:Method}
\begin{figure}[t]
    \centering
    \includegraphics[width=0.66\textwidth]{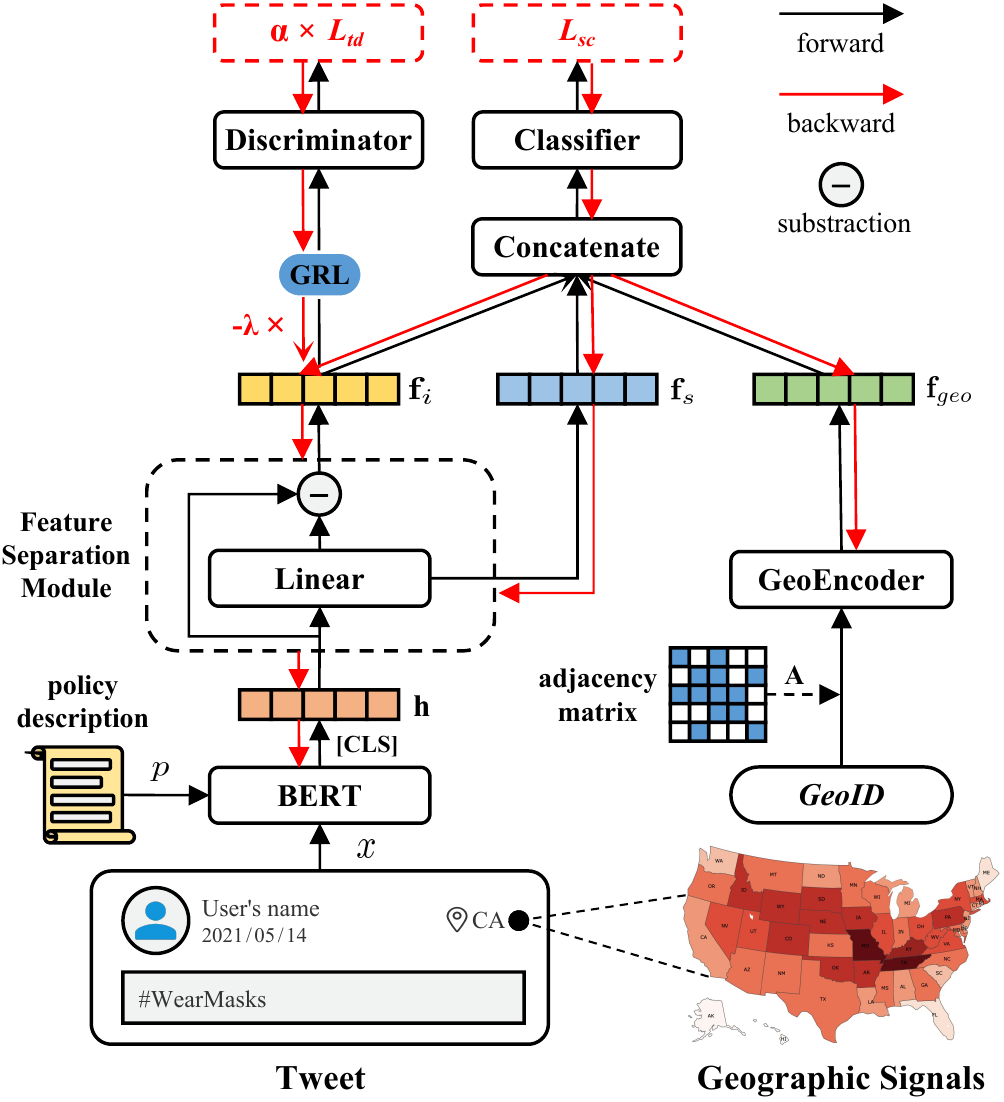}
    \caption{Overall architecture of proposed model.}
    \label{fig:model}
\end{figure}
\noindent \textbf{Problem formulation.} Given a set of labeled texts from the source target(s) $D_s=\{(t_{s}^i,x_{s}^i,y_{s}^i)\}_{{i=1}}^{N_s}$ and a set of unlabeled texts from a destination target (i.e., unseen target) $D_d=\{(t_{d}^i,x_{d}^i)\}_{i=1}^{N_d}$, where $t$ is the target involved in the text $x$, and $y$ is the corresponding stance label. We also have a large set of unlabeled texts from both targets $D_u=\{(t_{u}^i,x_{u}^i)\}_{i=1}^{N_u}$ (called external data) for adversarial training, where each sample belongs to the source target(s) or the destination target. Cross-target stance detection task is using one source target to predict stance labels (i.e., ``\textit{favor}", ``\textit{against}", or ``\textit{none}") of texts in $D_d$, while zero-shot setting is using multiple source targets. Note that, each sample in $D_s$ or $D_d$ is associated with a Geographic Identifier ($GeoID$) indicating the user's location. In Fig. \ref{fig:model}, we illustrate an overview architecture of our proposed model.

\subsection{Encoder Module}
\noindent BERT \cite{devlin2018bert} has shown tremendous success in various Natural Language Processing (NLP) tasks. To take full advantage of contextual information, we could jointly condition and embed topic $t$ and text $x$ using BERT. However, the original topic words are too short, which results in limited targeted background knowledge available for the stance classifier. Since the proposal of health policies is accompanied by the corresponding descriptions and people's attitudes are based on the contents or even the details of the policies, instead of using the topic words directly, we leverage external knowledge about the topic (i.e., policy descriptions) to enhance the model’s deeper understanding of health policies. We tokenize the text $x$ and policy-related descriptions $p$ as input sentence pair:
\begin{equation}\label{eq:bert}
    \textbf{h}=\text{BERT}([CLS]p[SEP]x[SEP])_{[CLS]},
\end{equation}

\noindent where $[CLS]$ is a special symbol that be used as the aggregate representation for the overall semantic context, while $[SEP]$ is the special separator token. $\textbf{h}$ is the hidden state of $[CLS]$ in the final layer and we use it for subsequent modeling.

\subsection{Feature Separation Module}
\noindent The contextual representation generated by BERT contains both topic-specific and topic-invariant information. To allow the model to generalize to unseen topics, it is essential and effective to learn and utilize transferable topic knowledge (i.e., topic-invariant information). Inspired by \cite{cui2020gradually}, we employ a simple linear transformation to separate and distinguish topic-specific and topic-invariant information, which can reduce the transfer difficulty without removing stance cues. First, we employ a linear layer to extract topic-specific features $\textbf{f}_{s}$:
\begin{equation}\label{eq:specific}
    \textbf{f}_{s}=\textbf{W}_{fs}\textbf{h}+\textbf{b}_{fs},
\end{equation}

\noindent where $\textbf{W}_{fs}$ and $\textbf{b}_{fs}$ are weight parameters. By removing target-specific features from the $\textbf{h}$, results in the constructed target-invariant representation $\textbf{f}_{i}$:
\begin{equation}
    \textbf{f}_{i}=\textbf{h}-\textbf{f}_{s}.
\end{equation}

The target-invariant features will enable the transferability of the model and serve as the input of the stance classifier and topic discriminator.

\subsection{GeoEncoder Module}
\noindent Geographic signals are spatial information, which can reflect potential characteristics and profiles of groups. Infusing geographic signals can provide region-specific features for model learning since people's attitudes across different regions are influenced by regional background factors, such as cultural background, political ideologies, and epidemiological context. Besides, graph structures (e.g., geographical topology) that describe the connectivity among regions grant us to explore underlying relationships between different regions and also can reflect some unobserved background factors. Inspired by this, we propose a GeoEncoder which utilizes geographic signals to capture the hidden background factors as non-text features $\mathbf{f}_{geo}$. Given the $GeoID$ of a text and the geographical adjacency matrix $\textbf{A}$\footnote{By default, each region is adjacent to itself.}, we leverage Graph Convolution Network (GCN) \cite{kipf2016semi} to make regions aggregate information from their neighboring regions:
\begin{equation}\label{eq:gcn}
    \textbf{E}^{(l)}=\sigma(\textbf{A}\textbf{E}^{(l-1)}\textbf{W}^{(l-1)}),
\end{equation}

\noindent where $\textbf{W}^{(l)}$ is a layer-specific weight matrix, $\sigma(\cdot)$ is ReLU activation function, and $\textbf{E}^{(l)}$ is the embedding matrix at $l$-th layer, with $\textbf{E}^{(0)}=\textbf{E}$. $\textbf{E}\in\mathbb{R}^{N\times{F}}$ are learnable embedding vectors specified by the $GeoIDs$ and $\textbf{f}_{geo}=\textbf{E}^{(l)}_{[GeoID]}$, $N$ means the number of total regions, $F$ is the hidden dimension of $\mathbf{f}_{geo}$.


\subsection{Stance Classifier}
\noindent  We apply a linear layer with softmax as the stance classifier to predict the stance labels of texts. We combine three features ($\textbf{f}_i$, $\textbf{f}_s$, and $\textbf{f}_{geo}$) to obtain the final representation for joint modeling. We use Cross-Entropy loss as loss function:
\begin{equation}
    \hat{y}_{sc}=\text{Softmax}(\textbf{W}_{sc}(\textbf{f}_i \oplus \textbf{f}_s \oplus \textbf{f}_{geo})+\textbf{b}_{sc}),
\end{equation}
\begin{equation}
    \mathcal{L}_{sc}=\sum_{x\in D_s}\text{CrossEntropy}(y_{sc},\hat{y}_{sc}),
\end{equation}

\noindent where $y_{sc}$ is the ground truth stance label, $\textbf{W}_{sc}$ and $\textbf{b}_{sc}$ are parameters of stance classifier, and $\oplus$ represents the concatenation operation.

\subsection{Topic Discriminator}
\noindent To further ensure the topic-invariant representation can distill more transferable topics knowledge to facilitate model adaptation across different topics, inspired by \cite{wei2019modeling,allaway2021adversarial}, we use a linear network with softmax as a topic discriminator to classify the corresponding topic label based on topic-invariant features. The topic discriminator is also trained by minimizing the Cross-Entropy loss:
\begin{equation}
    \hat{y}_{td}=\text{Softmax}(\textbf{W}_{td}\textbf{f}_i+\textbf{b}_{td}),
\end{equation}
\begin{equation}
    \mathcal{L}_{td}=\sum_{x\in D_s\cup D_u}\text{CrossEntropy}(y_{td},\hat{y}_{td}),
\end{equation}

\noindent where $y_{td}$ is the ground truth topic label, and $\textbf{W}_{td}$ and $\textbf{b}_{td}$ are parameters of topic discriminator. The training process of discriminator and topic-invariant features $\textbf{f}_{i}$ is adversarial. Specifically, target-invariant features aim to become generic enough to confuse topic discriminator (maximize $\mathcal{L}_{td}$) while topic discriminator makes efforts to correctly classify the topic labels (minimize $\mathcal{L}_{td}$). Following \cite{ganin2015unsupervised}, we adopt the gradient reversal layer (GRL) in our model, which is a widely used technique in transfer learning-based methods. The adversarial training process is essentially a minmax game as follows:
\begin{equation}
    \underset{\Theta_{\text{M}}}{\text{min}} \underset{\textbf{W}_{td},\textbf{b}_{td}}{\text{max}} \mathcal{L}_{sc}-\alpha\mathcal{L}_{td},
\end{equation}


\noindent where $\Theta_{\text{M}}$ including fine-tunable $\text{BERT}$, GeoEncoder, $\textbf{W}_{fs}$, $\textbf{b}_{fs}$, $\textbf{W}_{sc}$, and $\textbf{b}_{sc}$, and $\alpha$ is a trade-off parameter. During the forward propagation, GRL acts as an identity transform. During the back-propagation though, GRL takes the gradient from the subsequent level, multiplies it by $-\lambda$, and passes it to the preceding layer for more stable training and update \cite{ganin2015unsupervised,wei2019modeling}.

\begin{table}[t]
    \renewcommand{\arraystretch}{1.4} 
    \centering
    \scriptsize
    \caption{Dataset summary.}
    \begin{threeparttable}
    \setlength{\tabcolsep}{2.2pt}{
    \begin{tabular}{|cccc|}
        \hline
        \textbf{Topic} & \textbf{\#Unlabeled} & \tabincell{c}{\textbf{\#Labeled} \\ \textbf{\scriptsize{(Favor/Against/None)}}} & \textbf{Keywords}\tnote{*} \\
        \hline
        Stay at Home (SH) & 778 & 420 (194/113/113) & \tabincell{c}{lockdown, stayhome} \\
        Wear Masks (WM) & 1030 & 756 (173/288/295) & \tabincell{c}{mask, facemasks}\\
        Vaccination (VA) & 1535 & 526 (106/194/226) & \tabincell{c}{vaccine, vaccination} \\
        \hline
    \end{tabular}
    }
    \begin{tablenotes}
        \footnotesize
        \item[*] For more keywords of crawler, please refer to the source code\textsuperscript{\ref{code}}. 
      \end{tablenotes}
    \end{threeparttable}
    \label{tab:dataset}
\end{table}

\section{Experiments}\label{sec:ES}
\subsection{Experimental Setup}

\noindent \textbf{Data.} Previous studies \cite{glandt2021stance,cotfas2021longest} have built COVID-19-related stance datasets, where tweets are manually annotated as three stance labels: \textit{favor}, \textit{against}, and \textit{none} (i.e., neutral). In this work, we follow their research and adopt three health policies: (\uppercase\expandafter{\romannumeral1}) Stay at Home Order, (\uppercase\expandafter{\romannumeral2}) Wear Masks and (\uppercase\expandafter{\romannumeral3}) Vaccination, and we only select labeled tweets posted in the USA. Meanwhile, we also collect unlabeled tweets for these three policies via Twitter API\footnote{https://developer.twitter.com/}. The statistics of prepared data are summarized in Table \ref{tab:dataset}. Due to space limitations, please refer to our source code for detailed descriptions of the dataset (policy descriptions, keywords, etc.). Codes and other resources are publicly available.\footnote{https://github.com/Xiefeng69/stance-detection-for-covid19-related-health-policies\label{code}}

\begin{table}[t]
    \renewcommand{\arraystretch}{1.3} 
    \centering
    \scriptsize
    \caption{Performance comparison of cross-target stance detection measured by $F_{avg}$ and $F_m$. \textbf{Bold face} indicates the best result of each column and \underline{underlined} the second-best. Relative gain is compared with the second best result.}
    \setlength{\tabcolsep}{3.6pt}{
    \begin{tabular}{|l|cccccccccccc|}
         \hline
         \multirow{3}{*}{\textbf{Models}}& \multicolumn{12}{c|}{\textbf{Cross-target settings (\%)}}\\
         \cline{2-13}
         & \multicolumn{2}{c}{\textbf{SH$\rightarrow$WM}} & \multicolumn{2}{c}{\textbf{SH$\rightarrow$VA}} & \multicolumn{2}{c}{\textbf{WM$\rightarrow$SH}} & \multicolumn{2}{c}{\textbf{WM$\rightarrow$VA}} & \multicolumn{2}{c}{\textbf{VA$\rightarrow$SH}} & \multicolumn{2}{c|}{\textbf{VA$\rightarrow$WM}}\\
         \cline{2-13}
         & $F_{avg}$ & $F_{m}$ & $F_{avg}$ & $F_{m}$ & $F_{avg}$ & $F_{m}$ & $F_{avg}$ & $F_{m}$ & $F_{avg}$ & $F_{m}$ & $F_{avg}$ & $F_{m}$\\
        \hline
        BiLSTM & 25.4 & 30.6 & 25.6 & 30.5 & 39.1 & 45.1 & 40.8 & 47.5 & 31.5 & 38.1 & 33.0 & 38.6\\
        BiCond & 29.0 & 33.1 & 30.1 & 34.5 & 37.3 & 42.1 & 37.5 & 44.4 & 33.8 & 40.2 & 35.5 & 40.9\\
        TextCNN & 34.6 & 37.8 & 31.5 & 36.6 & 39.4 & 43.9 & 37.6 & 42.5 & 30.7 & 33.3 & 35.8 & 38.5\\ 
        TAN & 44.3 & 46.2 & 34.5 & 39.0 & 45.5 & 47.4 & 45.1 & 48.5 & 37.7 & 38.2 & 42.6 & 44.1\\ 
        CrossNet & \underline{45.7} & \underline{49.9} & \underline{39.4} & 43.6 & 43.4 & 47.3 & 47.7 & 50.7 & 37.7 & 38.3 & 46.7 & 48.1\\ 
        BERT & 44.7 & 49.3 & 34.9 & 41.2 & 44.3 & 49.7 & 52.6 & 55.3 & \underline{44.4} & \underline{45.6} & \underline{53.7} & \underline{55.1}\\ 
        WS-BERT-S & 45.4 & 49.1 & \textbf{40.3} & \textbf{44.9} & 41.9 & 48.0 & 51.0 & 54.8 & 39.9 & 41.4 & 47.2 & 49.9\\ 
        WS-BERT-D & 40.1 & 47.1 & 30.5 & 38.9 & \underline{48.2} & \underline{52.5} & \underline{55.4} & \underline{57.5} & 43.5 & 44.9 & 49.5 & 51.1\\ 
        \hline
        \rowcolor{gray!20} Ours & \textbf{47.6} & \textbf{51.9} & \underline{39.4} & \underline{44.4} & \textbf{50.9} & \textbf{54.1} & \textbf{57.6} & \textbf{59.3} & \textbf{46.1} & \textbf{47.4} & \textbf{54.5} & \textbf{56.3} \\ 
        Improve (\%) & 4.2\% & 4.0\% & - & - & 5.6\% & 3.0\% & 3.9\% & 3.1\% & 3.8\% & 3.9\% & 1.5\% & 2.1\%\\ 
        \hline
    \end{tabular}}
    \label{tab:performace-comparison-cross}
\end{table}

\medskip
\noindent \textbf{Baselines and Evaluation Metrics.} We select the following methods in the literature as comparison baselines: (1) neural network-based methods \texttt{BiLSTM} \cite{hochreiter1997long}, \texttt{BiCond} \cite{augenstein2016stance}, and \texttt{TextCNN} \cite{chen2015convolutional}; (2) attention-based methods \texttt{TAN} \cite{du2017stance} and \texttt{CrossNet} \cite{xu2018cross}; (3) BERT-based methods \texttt{BERT} \cite{devlin2018bert} and \texttt{WS-BERT} \cite{he2022infusing}. Besides, we adopt \texttt{WS-BERT-S} and \texttt{WS-BERT-D} which encode Wikipedia knowledge in the single manner and dual manner, respectively \cite{he2022infusing}. Following previous works \cite{allaway2021adversarial,xu2018cross}, we utilize the average F1-score (denoted as $F_{avg}$) and the average of both micro-average and macro-average F1-scores (denoted as $F_m$) as evaluation metrics.

\noindent \textbf{Implementation Details.} All the experiments are performed in \textit{Pytorch} on \textit{NVIDIA GeForce 3090 GPU}. The reported results are the averaged score of 5 runs with different random initialization. In cross-target setting, the models are trained and validated on one topic and evaluated on another. There can be categorized into six source$\rightarrow$destination tasks for cross-target evaluation: \textbf{SH$\rightarrow$WM}, \textbf{SH$\rightarrow$VA}, \textbf{WM$\rightarrow$SH}, \textbf{WM$\rightarrow$VA}, \textbf{VA$\rightarrow$SH}, and \textbf{VA$\rightarrow$WM}. In zero-shot setting, the models are trained and validated on multiple topics and tested on one unseen topic. We use the unseen topic's name as the task's name, thus, the zero-shot evaluation can be set into: \textbf{SH}, \textbf{WM}, and \textbf{VA}. For all tasks, the batch size is set to 16, the dropout rate is set to 0.1, and the input texts are truncated or padded to a maximum of 100 tokens. We train all models using AdamW optimizer with weight decay 5e-5 for a maximum of 100 epochs with patience of 10 epochs, and the learning rate is chosen in \{1e-5, 2e-5\}. In our model, we adopt the pre-trained uncased BERT-base as the encoder. The maximum length of policy descriptions is fixed at 50, the layer number $l$ of GCN is set to 2, the trade-off parameter $\alpha$ is set to 0.01, the GRL's parameter $\lambda$ is set to 0.1, and the hidden dimension of GeoEncoder is optimized in \{128, 256\}. Please refer to our project repository\textsuperscript{\ref{code}} for more experimental implementation details (hyper-parameter settings, training details, data preparation, etc.).

\begin{table*}[t]
    \renewcommand{\arraystretch}{1.3} 
    \centering
    \scriptsize
    \caption{Performance comparison of zero-shot stance detection.}
    \setlength{\tabcolsep}{3.6pt}{
    \begin{tabular}{|l|cccccc|}
         \hline
         \multirow{3}{*}{\textbf{Models}} & \multicolumn{6}{c|}{\textbf{Zero-shot settings (\%)}}\\
         \cline{2-7}
         & \multicolumn{2}{c}{\textbf{SH}} & \multicolumn{2}{c}{\textbf{WM}} & \multicolumn{2}{c|}{\textbf{VA}} \\
         \cline{2-7}
         & $F_{avg}$ & $F_{m}$ & $F_{avg}$ & $F_{m}$ & $F_{avg}$ & $F_{m}$ \\
        \hline
        BiLSTM & 45.6 & 49.6 & 25.7 & 31.7 & 36.7 & 42.4 \\
        BiCond & 45.7 & 50.1 & 29.8 & 34.6 & 29.6 & 35.2 \\ 
        TextCNN & 41.0 & 41.5 & 35.7 & 37.8 & 34.8 & 39.2 \\ 
        TAN & 45.8 & 47.7 & 50.2 & 51.7 & 46.5 & 49.3 \\ 
        CrossNet & 45.9 & 49.3 & 55.6 & 56.8 & 45.1 & 48.2 \\ 
        BERT & 49.6 & 53.8 & \underline{63.4} & \underline{64.6} & \underline{57.5} & \underline{59.4} \\ 
        WS-BERT-S & 48.6 & 53.0 & 61.0 & 62.4 & 55.6 & 57.9 \\ 
        WS-BERT-D & \underline{51.6} & \underline{55.2} & 61.6 & 63.3 & 55.3 & 57.6 \\ 
        \hline
        \rowcolor{gray!20} Ours & \textbf{53.3} & \textbf{56.2} & \textbf{65.1} & \textbf{66.4} & \textbf{58.9} & \textbf{59.9} \\ 
        Improve (\%) & 3.3\% & 1.8\% & 2.7\% & 2.8\% & 2.4\% & 0.8\% \\ 
        \hline
    \end{tabular}}
    \label{tab:performace-comparison-zero}
\end{table*}

\subsection{Experimental Results}
\noindent We evaluate all models both in cross-target and zero-shot settings, and the results are reported in Table \ref{tab:performace-comparison-cross} and Table \ref{tab:performace-comparison-zero}, respectively. There is an overall phenomenon that the accuracy in zero-shot setting is better than cross-target setting. This is because zero-shot stance detection leverages a broader source of supervision, which allows models have better semantic understanding across different topics. Our proposed method outperforms comparison baselines on most tasks and improves the average $F_{avg}$ and $F_{m}$ by 3.5\% and 2.7\%, respectively.

We observe that \texttt{BiLSTM}, \texttt{BiCond}, and \texttt{TextCNN} overall perform worst mainly because they do not explicitly use topic information or have limited perception of unknown destination topic, which affirms the importance of designing a topic-aware approach for cross-target and zero-shot settings. For \texttt{TAN} and \texttt{CrossNet}, by designing an attention mechanism in the network to notice the importance of each word and reflect users' concerns, they can provide interpretable evidence to enable semantic understanding across different topics. \texttt{BERT} is still a comprehensive baseline, even though it ignores transferable knowledge between topics, but has strong generalization ability because it learns from a large-scale unsupervised corpus. Therefore, \texttt{BERT} exploits rich semantic information to perform relatively good performance in both scenarios. \texttt{WS-BERT} is a recent powerful stance classifier that leverages external knowledge from Wikipedia as a bridge to enable model's deeper understanding and then precisely captures the stance towards a topic. However, since \texttt{WS-BERT} does not make constraints for learning transferable knowledge between different topics in the training phase, it will prone to fitting to a specific topic, which results in bringing performance drops when topic transfer. Compared with the above baselines, our proposed model takes topic information, external knowledge (i.e., policy descriptions), and non-text information (i.e., geographic signals) into account to improve model's discriminability. Meanwhile, adversarial learning is applied as a constraint to learn topic-invariance to facilitate enhancing the model's transferability.

\subsection{Model Size and Efficiency}

\begin{table}[t]
    \renewcommand\arraystretch{1.3}
    \scriptsize
    \caption{Comparison of model size and efficiency of training and testing on SH.}
    \centering
    \begin{tabular}{|l|ccc|}
        \hline
        \textbf{Models} & \tabincell{c}{\textbf{Parameters}\\(\textit{Million})} & \tabincell{c}{\textbf{Training Time}\\(\textit{s}/\textit{epoch})} & \tabincell{c}{\textbf{Inference Time}\\(\textit{s})} \\
        \hline
        BiLSTM & 39.6 & 2.6 & 0.21 \\
        BiCond & 41.5 & 2.9 & 0.23\\
        TextCNN & 40.4 & 1.1 & 0.03\\
        TAN & 40.2 & 2.3 & 0.24\\
        CrossNet & 41.6 & 4.1 & 0.24\\
        BERT & 109.5 & 11.2 & 0.61\\
        WS-BERT-S & 218.9 & 13.7 & 1.01\\
        WS-BERT-D & 218.9 & 15.6 & 1.28\\
        \hline
        ours & 110.1 & 23.7 & 0.86\\
        \hline
    \end{tabular}
    \label{tab:modelsize}
\end{table}

\begin{figure}[b]
    \centering
    \includegraphics[width=0.65\linewidth]{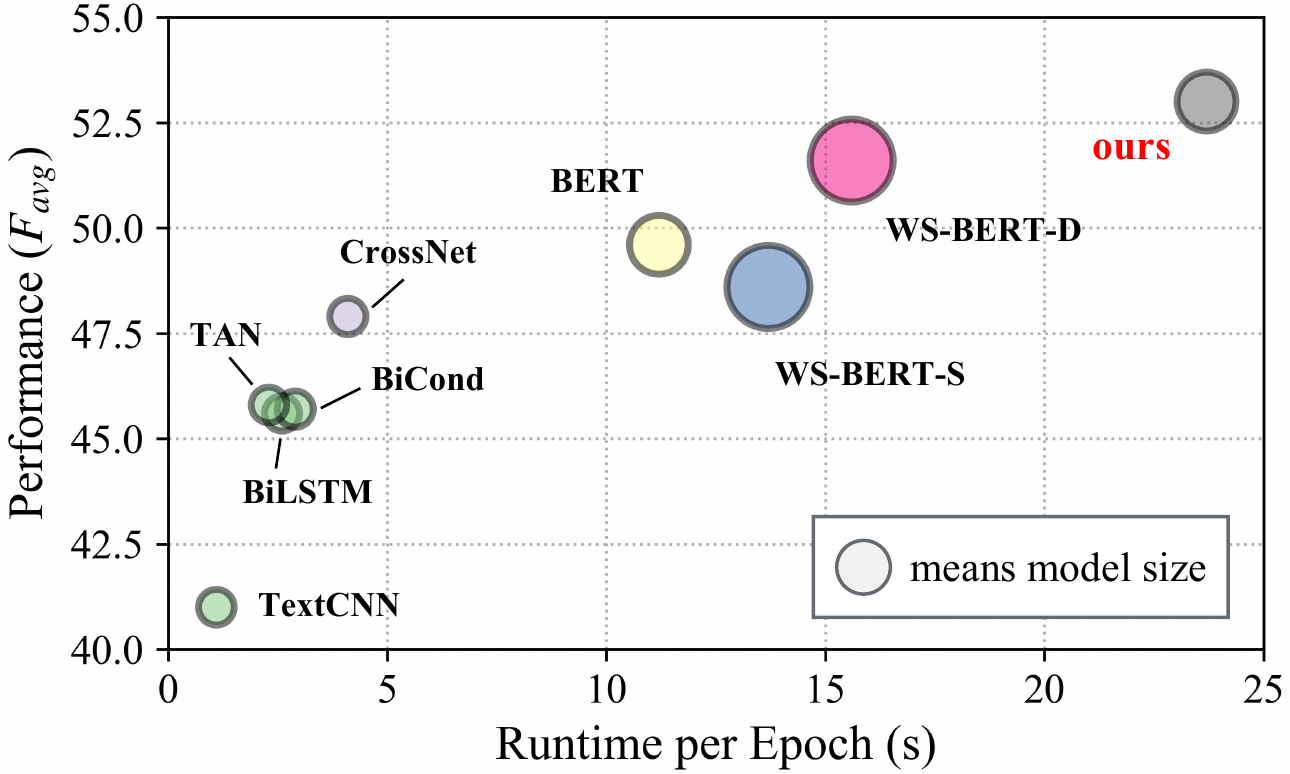}
    \caption{Model size and efficiency comparison on SH task.}
    \label{fig:modelsize}
\end{figure}

\noindent In Table \ref{tab:modelsize}, we record and compare the model size and efficiency of all methods on \textbf{SH} task, which uses the most samples for training. For the legibility of the comparison results, we plot the relationship between model performance, model size, and training time in Fig. \ref{fig:modelsize}. BERT-based models have large trainable parameters because increasing model size when pretraining natural language representations often result in improved performance on downstream tasks. Compared to \texttt{WS-BERT}, our proposed model has an acceptable model size. Among all models, \texttt{TextCNN} is the fastest in terms of training time and inference time, as it applies convolutional neural networks to capture word coherence in parallel. Our model has the longest training time per epoch, this is because, during the adversarial learning process, it is necessary to leverage a large amount of unlabeled data to train the discriminator for domain adaptation. Therefore, the increment of training samples will inevitably increase the training time per epoch. Despite this, adversarial learning has no obvious adverse effect on inference time. Thus, the increase in training time can be seen as a trade-off against the performance gains obtained by training the discriminator to better extract transferable knowledge.

\section{Conclusion and Future Work}
\noindent In this paper, we consider a  more realistic scenario on stance detection (i.e., cross-target and zero-shot settings) for the pandemic and propose a novel and practical stance classifier to automatically identify the public’s attitudes toward COVID-19-related health policies. Specifically, we apply adversarial learning to ensure model's transferability to emerging policies. Moreover, to enhance the model's deeper understanding, we infuse policy descriptions as external knowledge and devise a GeoEncoder to capture regional background factors as non-text features. Extensive experiments demonstrate that our proposed method achieves state-of-the-art performance in both cross-target and zero-shot settings.

In the future, we will devote to building a high-accuracy stance detection application for public health policies towards COVID-19 by utilizing multi-aspect information (e.g., stance expressions, sentiments, and epidemic trends).
\medskip

\noindent{\textbf{Acknowledgments.}} We thank reviewers for their helpful feedback. This work is supported by the National Natural Science Foundation of China No. 62172428.

\bibliographystyle{splncs04}
\bibliography{reference}

\end{document}